\documentclass[preprint]{nesy2025} 

\usepackage{wrapfig}
\usepackage{multicol}
\usepackage{soul} 
\usepackage[T1]{fontenc}
\usepackage{lineno} 
\usepackage[utf8]{inputenc}

\usepackage{microtype}

\usepackage{inconsolata}

\usepackage{amsmath}

\usepackage{multirow}
\usepackage{tcolorbox}
\tcbuselibrary{breakable, skins}

\usepackage{comment}
\newcommand{\mee}[1]{\textcolor{red}{#1}}

\usepackage{booktabs}
 \usepackage{siunitx}

\theorembodyfont{\upshape}
\theoremheaderfont{\scshape}
\theorempostheader{:}
\theoremsep{\newline}

\title[Enhancing LLMs with Neurosymbolic Reasoning for Multilingual Tasks]{Enhancing Large Language Models with Neurosymbolic Reasoning for Multilingual Tasks}

  \clearauthor{\Name{Sina {Bagheri Nezhad}}
  \Email{sina.bagherinezhad@pdx.edu}
  \\
   \Name{Ameeta Agrawal}
   \Email{ameeta@pdx.edu}\\
   \addr Portland State University, USA}

\begin{document}
\maketitle

\thispagestyle{empty} 

\begin{abstract}
Large language models (LLMs) often struggle to perform multi-target reasoning in long-context scenarios where relevant information is scattered across extensive documents. To address this challenge, we introduce {NeuroSymbolic Augmented Reasoning (NSAR)}, which combines the benefits of neural and symbolic reasoning during inference. NSAR explicitly extracts symbolic facts from text and generates executable Python code to handle complex reasoning steps. Through extensive experiments across seven languages and diverse context lengths, we demonstrate that NSAR significantly outperforms both a vanilla RAG baseline and advanced prompting strategies in accurately identifying and synthesizing multiple pieces of information. Our results highlight the effectiveness of combining explicit symbolic operations with neural inference for robust, interpretable, and scalable reasoning in multilingual settings.
\end{abstract}

\section{Introduction}

Large language models (LLMs) have achieved impressive progress in natural language processing, yet their ability to handle \textit{long-context, cross-lingual reasoning} remains limited \citep{agrawal2024evaluatingmultilinguallongcontextmodels}. In real-world multilingual scenarios, information is often scattered across languages and embedded within lengthy documents, complicating retrieval and coherent reasoning. This challenge is exacerbated by the \textit{lost-in-the-middle} phenomenon, where LLMs struggle to retain and integrate relevant information from extended contexts \citep{liu2023lostmiddlelanguagemodels}. Although retrieval-augmented generation (RAG) frameworks help narrow the context window, LLMs still frequently fail to combine dispersed facts into a consistent chain of reasoning, resulting in hallucinations and inconsistencies \citep{zhu2024largelanguagemodelsinformation, jiang2024longragenhancingretrievalaugmentedgeneration}.

In this work, we introduce \textbf{NeuroSymbolic Augmented Reasoning (NSAR)}—a neurosymbolic method that explicitly merges symbolic reasoning with neural inference. NSAR centers on a \textit{neurosymbolic prompt} that instructs LLMs to extract structured symbolic facts and generate executable code, thus offering a verifiable and interpretable reasoning pathway for complex tasks. By integrating symbolic logic into a retrieval-based framework, NSAR improves interpretability, consistency, and accuracy when working with multilingual and long-context data. 

The main contributions of our work are as follows:
\begin{itemize}
    \item We propose {NSAR}, a neurosymbolic reasoning framework that couples retrieval with explicit symbolic reasoning. We develop a dedicated {NSAR prompt} to structure information into symbolic representations and guide the model in generating executable Python code for final verification.
    \item We show that combining \textit{Chain of Thought (CoT), ReAct, and Self-Reflection} prompting strategies with neurosymbolic reasoning leads to further performance gains.
    \item Through extensive experiments on cross-lingual, long-context tasks, we demonstrate that NSAR substantially outperforms purely retrieval-based and neural-only methods.
\end{itemize}

\section{Related Works}

\paragraph{Long-Context and Cross-Lingual Challenges}
Large language models (LLMs) frequently exhibit a \textit{lost-in-the-middle} phenomenon in extended contexts \citep{liu2023lostmiddlelanguagemodels, xu2024retrieval}, causing them to overlook crucial information. Retrieval-augmented frameworks such as LONGEMBED \citep{zhu2024largelanguagemodelsinformation}, LongRAG \citep{jiang2024longragenhancingretrievalaugmentedgeneration} and DR-RAG \citep{hei2024drragapplyingdynamicdocument} alleviate this issue but mostly target monolingual settings. McCrolin \citep{limkonchotiwat-etal-2024-mccrolin} extends these ideas to cross-lingual retrieval, yet it remains computationally intensive. Bridging queries and contexts in different languages is still difficult \citep{li2024bordirlinesdatasetevaluatingcrosslingual, hengle2024multilingualneedlehaystackinvestigating}. Methods such as OPTICAL \citep{Huang_2023} and XAMPLER \citep{lin2024xamplerlearningretrievecrosslingual} improve multilingual retrieval but face scalability limitations in very long contexts. Parameter-efficient solutions, such as Sparse Fine-Tuning Masks (SFTMs) \citep{litschko-etal-2022-parameter} and adapters, achieve strong multilingual retrieval while reducing resource overhead, but they do not fully address multi-target reasoning.

\paragraph{Advanced Prompting Strategies}
Prompting methods like Chain-of-Thought (CoT) \citep{wei2022chain}, ReAct \citep{yao2023react}, and Self-Reflection \citep{renze2024selfreflectionllmagentseffects} improve LLM reasoning by encouraging more explicit intermediate steps. The Tree-of-Thoughts (ToT) approach \citep{yao2023tree} further expands these pathways, enhancing structured reasoning. However, such methods rely on text-based explanations that may lack verifiability.

\paragraph{Neurosymbolic Reasoning}
Neurosymbolic approaches offer interpretability and robustness by coupling neural networks with symbolic logic. Hybrid systems like LINC \citep{olausson-etal-2023-linc} and PAL \citep{gao2023palprogramaidedlanguagemodels} produce executable code, improving transparency and accuracy. Similarly, incorporating symbolic modules into decision-making \citep{fang2024large} enhances reliability.


\section{System Architecture and Approach}

\begin{figure}[th]
    \centering
    \includegraphics[width=1\linewidth]{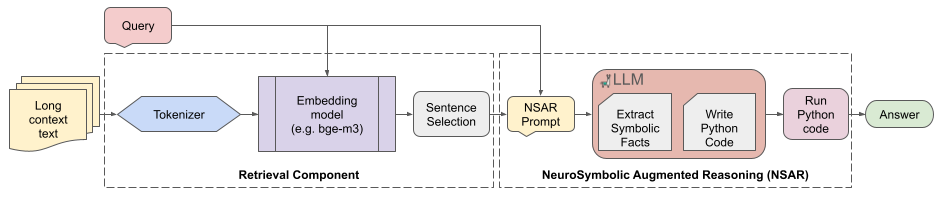}
    \caption{High-level overview of the system, illustrating the two layers: the \emph{retrieval component} (left) and the \emph{neurosymbolic reasoning component} (NSAR, right). First, the long context and query are tokenized and embedded to select the most relevant sentences. Next, the NSAR prompt directs the LLM to extract symbolic facts and generate Python code, which is then executed to produce the final answer.}
    \label{fig:overview} \vspace{-0.5cm}
\end{figure}

Our system, as visualized in Figure~\ref{fig:overview}, is composed of two distinct layers. First, a \emph{Retrieval Component} efficiently narrows down the long, multilingual context using a two-phase Retrieval-Augmented Generation (RAG) framework. Then, the \emph{NeuroSymbolic Augmented Reasoning (NSAR)} component builds upon this retrieved context to perform explicit neurosymbolic reasoning. NSAR employs a specialized prompt that instructs the language model to extract structured symbolic facts and generate executable Python code for deterministic reasoning.

\subsection{Retrieval Component}
The retrieval component extracts the most relevant chunks of information from extensive, multilingual contexts using a Retrieval-Augmented Generation framework.

\paragraph{Tokenization and Embedding}
The input context, which may comprise hundreds of thousands of words across multiple languages, is first segmented into sentences and embedded to capture semantic nuances. We use the Punkt tokenizer \citep{10.1162/coli.2006.32.4.485} for segmentation and the multilingual \textit{bge-m3} model \citep{chen-etal-2024-m3} to generate sentence embeddings in 1024-dimensional embedding size. 

\paragraph{Candidate Selection and Model Input}
Next, we compute the semantic distance between each sentence embedding and the query, selecting the top $k$ most relevant sentences based on a tunable hyperparameter (with $k$ values of 3, 5, 10, 20, and 50 evaluated in our experiments). These sentences are then passed to the language model as input for the subsequent neurosymbolic reasoning process.

\subsection{NeuroSymbolic Augmented Reasoning (NSAR)}

Purely neural approaches to long-context question answering often struggle with reliability, interpretability, and logically integrating multiple pieces of information. While recent prompting techniques (e.g., Chain-of-Thought, ReAct, and Self-Reflection) have improved reasoning, they remain reliant on implicit neural processes, leaving little room for explicit verification or modular correction. To address these limitations, we introduce the {NeuroSymbolic Augmented Reasoning (NSAR)} component which integrates structured symbolic representations within the neural architecture by extracting symbolic facts and generating executable Python code for reasoning. This approach bridges the gap between the \emph{flexibility and fluency} of LLMs and the \emph{interpretability and rigor} of symbolic methods. Neural systems excel at language understanding and generation, but often fail in complex scenarios requiring multiple reasoning steps, such as comparing multiple facts, deducing the “largest” or “smallest” value, or verifying constraints across scattered pieces of information. Symbolic methods, by contrast, excel in structured reasoning but can be brittle when parsing unstructured text. By coupling a language model with a symbolic layer, NSAR enhances interpretability by  providing an explicit record of extracted facts and logical steps, enabling users to audit and verify the model’s reasoning. Additionally, it improves reliability by reducing errors in compositional tasks by systematically comparing and fusing pieces of information through symbolic code execution. 

We design an \emph{NSAR prompt} to guide the reasoning process through three distinct stages:

\paragraph{1. Symbolic Fact Extraction} First, the model is instructed to identify all relevant facts in the provided context and represent them in a structured, symbolic format. For instance, if the context contains lines such as \textit{“The special magic Cairo number is: 1234567”} and \textit{“The special magic Mumbai number is: 9999999”}, the model generates:
    \begin{verbatim}
    FACT("Cairo", "special_magic_number", 1234567)
    FACT("Mumbai", "special_magic_number", 9999999)
    \end{verbatim}
    \vspace{-0.7cm}
    \paragraph{2. Python Code Generation} Next, the model is prompted to produce concise, executable Python code that uses the extracted symbolic facts to answer the question. Instead of implicitly inferring logic through text, the Python code can contain explicit comparisons (\verb|>, <, ==|), data structures (lists, dictionaries), or domain-specific libraries. In the case of identifying the \textit{largest special magic number}, this code might look like:
    \begin{verbatim}
    numbers = [1234567, 9999999]
    answer = max(numbers)
    \end{verbatim}
    \vspace{-0.5cm}
    The logic here can be arbitrarily extended to handle more complex reasoning steps (e.g., filtering facts, applying constraints, or computing aggregates). Once the LLM generates the Python code, it is executed in a controlled environment.
    \paragraph{3. Final Answer Extraction} Finally, the answer is determined by executing the generated Python code. This guarantees a concise, verified response and prevents any contradictory or incoherent rationales that might arise from purely text-based reasoning. In other words, while the LLM might propose a final textual answer, the \emph{actual} answer delivered to the user is the deterministic output of the code execution.

As such, NSAR prompt structure ensures that the language model provides an interpretable chain of reasoning, resulting in a Python snippet that can be independently executed and verified. The template of NSAR prompt is shown below:
\begin{tcolorbox}[title=NSAR Prompt Template, breakable, enhanced,fonttitle=\small]
\small
You are a helpful assistant that employs a neurosymbolic method. Given the following context and question, please follow these steps:

1. Extract all relevant facts from the context and represent them as symbolic facts using the format \texttt{FACT(entity, attribute, value)}.

2. Generate executable Python code that uses the extracted symbolic facts to compute the final answer.

3. Finally, output only the final answer.

\#CONTEXT \\
\{text\} \\
\#ENDCONTEXT

\#QUESTION \\
What is the largest special magic number?
\end{tcolorbox}

In this work, “neurosymbolic” denotes the hybrid of explicit fact extraction with deterministic code execution. While our current triples support only simple attribute logic, extending to richer formalisms (e.g. first-order rules or constraint solvers) would more fully realize the neurosymbolic ideal.

\section{Experiments}

\subsection{Dataset}
We evaluate the proposed approach on a question answering task encompassing long contexts and multiple languages. We adopt and extend the mLongRR dataset \citep{agrawal2024evaluatingmultilinguallongcontextmodels} by increasing the maximum context length to {512,000 words} and expanding the language set to {English, Vietnamese, Swahili, Persian, Russian, Hindi, and Arabic}. This setting incorporates multiple scripts (e.g., Cyrillic, Devanagari) and introduces a {cross-lingual} challenge: the {query is always in English} while the {context (haystack)} is in one of the seven languages. In addition, we randomly place {three “needles”} (target sentences) within each context haystack, requiring the model to retrieve and compare multiple pieces of relevant information.

The contexts consist of news articles ranging in size from {2k, 8k, 16k, 32k, 64k, 128k, 256k, and 512k words}. Embedded within these contexts are needles such as: {“The special magic \{city\} number is: \{number\}”}, where \{city\} is randomly chosen from 23 city names translated into all seven target languages, and \{number\} is a random 7-digit number \citep{agrawal2024evaluatingmultilinguallongcontextmodels, geminiteam2024gemini15unlockingmultimodal, TheC3}. Needles appear in the same language as the haystack, ensuring a consistent linguistic setting. Additional details on the needle translation and city selection process can be found in Appendix~\ref{sec:needels}.

\subsection{Evaluation}
To assess the effectiveness of NSAR in handling complex, cross-lingual reasoning tasks, we adopt an evaluation protocol based on a multi-target scenario—the 3-needles test. Previous studies have reported that the 3-needle scenario is particularly challenging for LLMs operating in long-context settings and low-resource languages \citep{agrawal2024evaluatingmultilinguallongcontextmodels}. In this test, three needles are randomly placed throughout the context, and the model is prompted with the query: “What is the largest special magic number?” This setup requires the model to locate and compare multiple pieces of information to produce the correct answer, reflecting real-world scenarios where critical information is scattered.

We run experiments across eight different context lengths (2k, 8k, 16k, 32k, 64k, 128k, 256k, and 512k words) and five values of $k$ for retrieved sentences (3, 5, 10, 20, and 50). For each setting, the three needles are randomly positioned in the text. 
We use accuracy as our main metric, defined as the percentage of cases in which the model correctly identifies the correct response by effectively integrating multiple retrieved facts. The results are reported in terms of average accuracy computed over two runs.

\subsection{Models and Baseline Methods}
\label{sec:models}

We evaluate our approach using two large language models. {GPT-4o-mini} is a smaller, resource-efficient variant of GPT-4o \citep{openai2024gpt4ocard}, designed to trade off some generative capacity for faster inference and lower memory usage, and {Llama 3.2 (90B)} which is an expanded successor to Llama 2 \citep{touvron2023llama2openfoundation}, with refined training procedures and a broader pretraining corpus.

We include three  relevant baseline prompting strategies like Chain-of-Thought \citep{wei2022chain}, ReAct \citep{yao2023react}, and Self-Reflection \citep{renze2024selfreflectionllmagentseffects} in our evaluation which have demonstrated success in boosting reasoning performance.


\begin{figure*}[!t]
  \centering
  \subfigure[Accuracy vs. Context Length]{
    \includegraphics[width=0.3\textwidth]{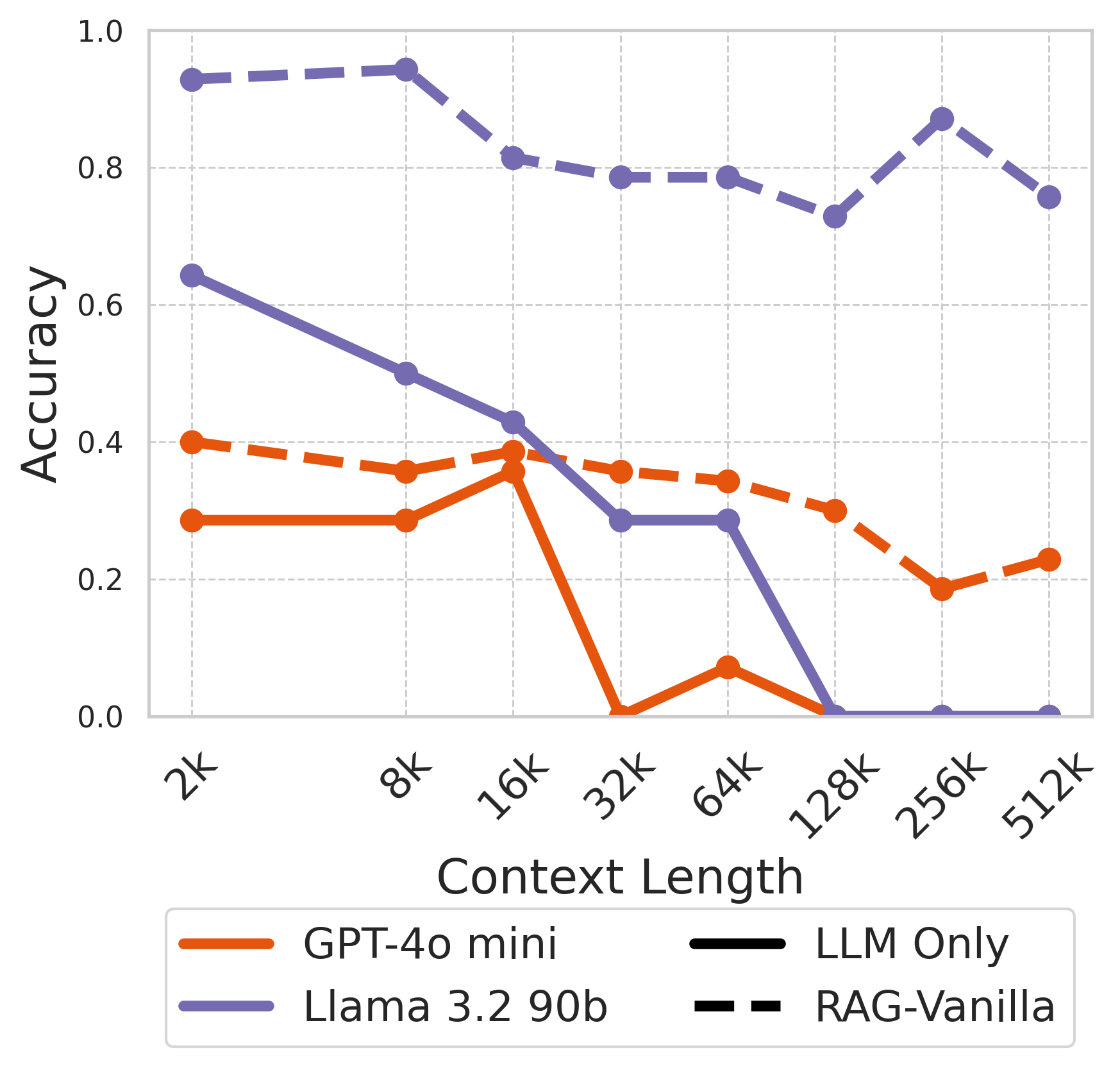}
    \label{fig:context_length_accuracy}
  }
  \quad \quad \quad \quad
  \subfigure[Radar Plots of Retrieval Accuracy]{
    \includegraphics[width=0.3\textwidth]{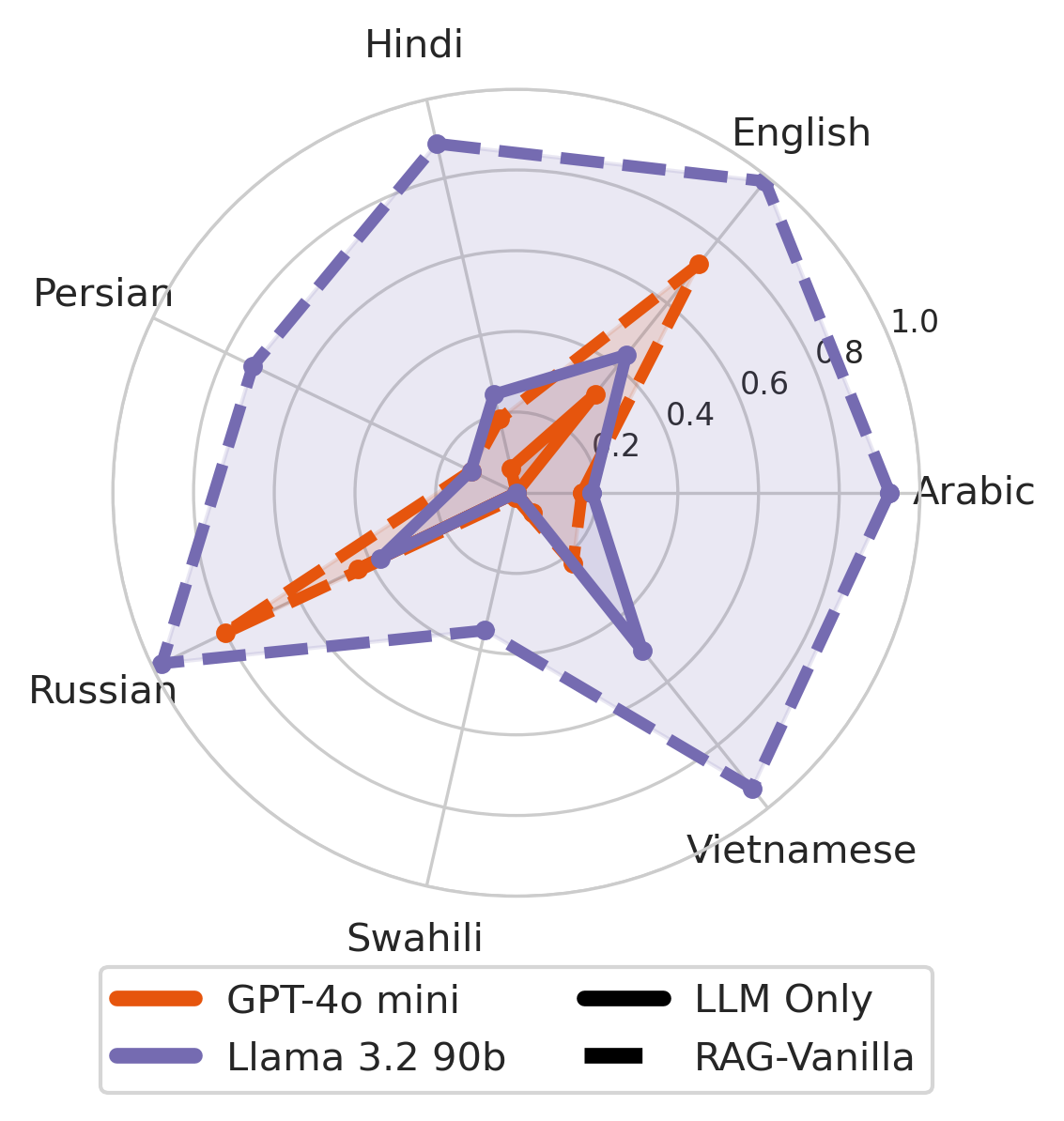}
    \label{fig:radar}
  }
  \caption{Accuracy as a function of context length and across seven context languages.}
  \label{fig:results} \vspace{-0.5cm}
\end{figure*}

\section{Results and Analysis}
\label{sec:results}


\subsection{LLM-Only {\em vs}. RAG-Vanilla}
We compare an LLM-only approach, where the entire context is fed directly to the model, with a RAG-Vanilla approach that employs the retrieval component to narrow the context (by identifying the top-$k$ most relevant sentences) before further processing with the LLM. This selective context narrowing not only improves retrieval accuracy but also significantly reduces computational costs by minimizing the total token count processed by the model. In both methods, we use the same prompt adapted from previous work (see Appendix~\ref{sec:prompts}). The performance results for the RAG-Vanilla approach shown in the figures represent the average accuracy across all k-values and evaluated languages.


\paragraph{Context length} From the results presented in Figure~\ref{fig:context_length_accuracy}, we observe that  both GPT-4o-mini and Llama 3.2 exhibit declining accuracy as context length increases in LLM only (solid lines) scenario, with accuracy dropping to almost 0\% as context lengths reach 128K words. In contrast, narrowing the input to a small set of top-relevant sentences using RAG-Vanilla (dashed lines) enables these models to maintain high accuracy even in long-context scenarios (over half a million words).

\paragraph{Accuracy by context language} Figure~\ref{fig:radar} shows radar plots comparing  accuracy across seven context languages using English prompts for the 3-Needles test. We observe that RAG-Vanilla consistently outperforms the LLM-only baseline across all tested languages, including those with non-Latin scripts such as Persian, Russian, Hindi, and Arabic. Notably, even languages like Swahili and Vietnamese, which differ significantly from English in morphology, benefit from the narrowed context, highlighting the method’s cross-lingual robustness. 

\begin{figure}
  \centering
  \subfigure[Retrieval failures (red) versus LLM failures (blue) for GPT-4o-mini and Llama 3.2.]{
    \includegraphics[width=0.45\textwidth]{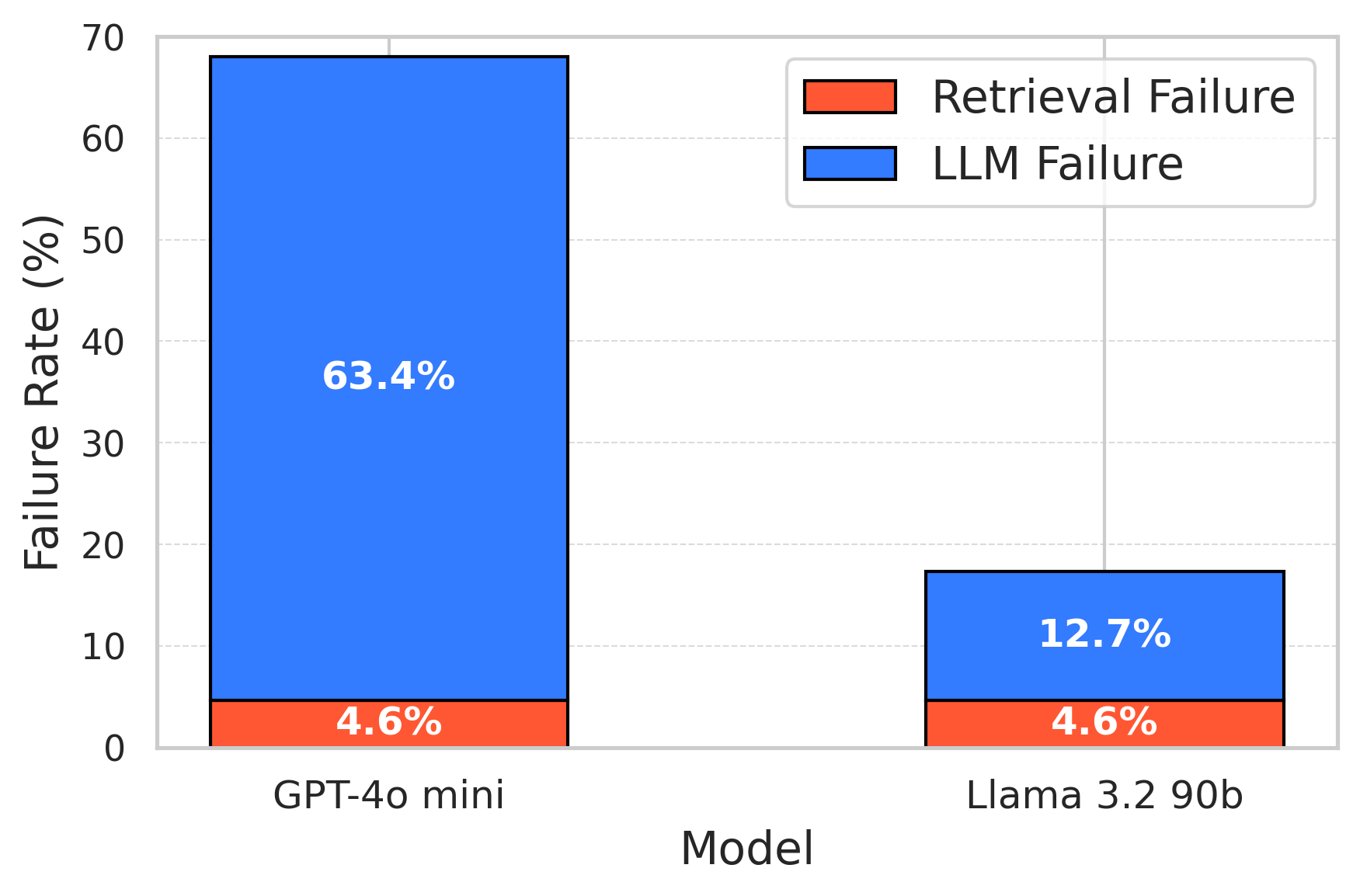}
    \label{fig:failure_rates}
  }
  \hfill
  \subfigure[Retrieval errors and LLM errors as $k$ (the number of retrieved sentences) increases.]{
    \includegraphics[width=0.45\textwidth]{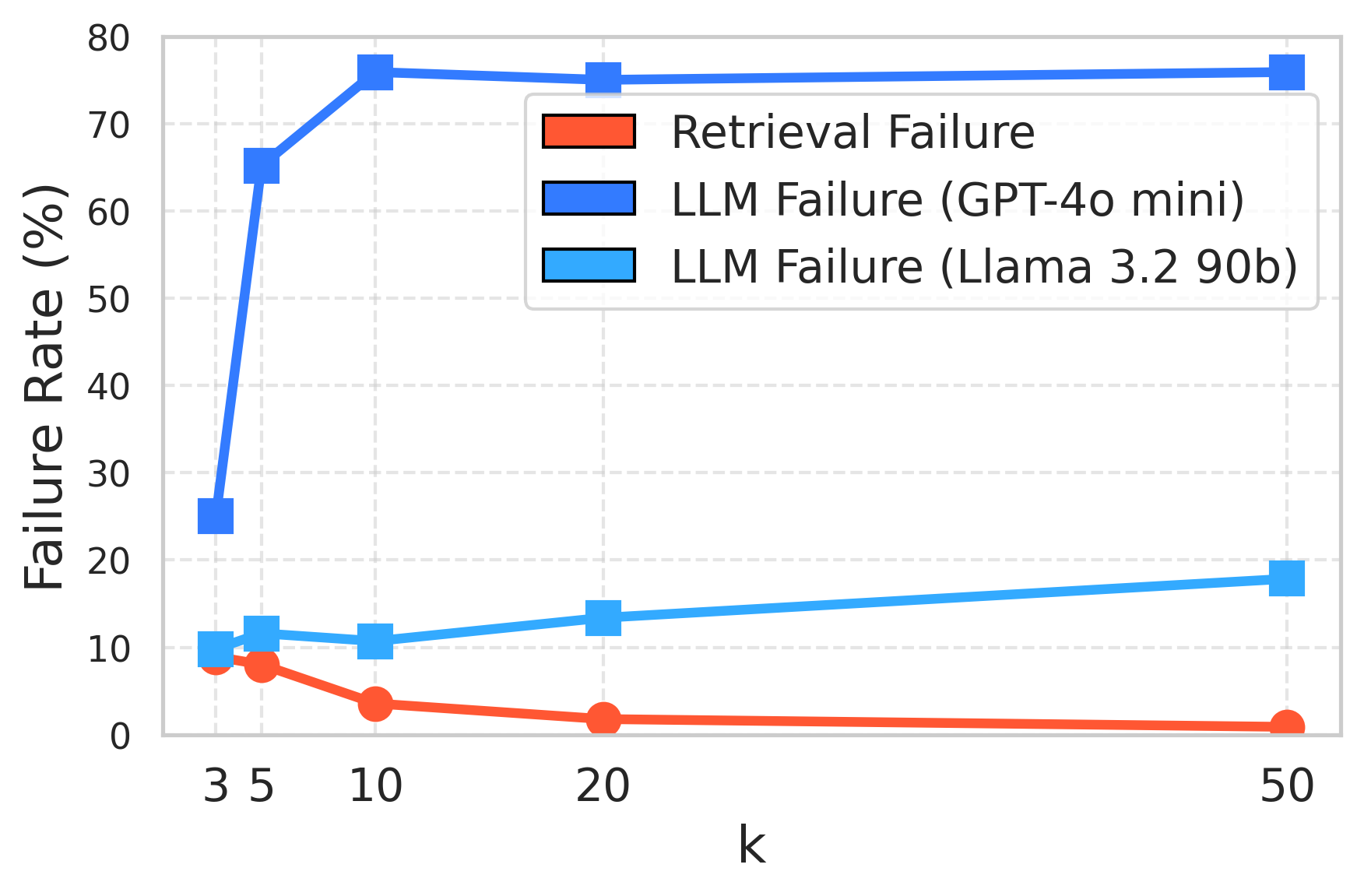}
    \label{fig:sentence_cap_analysis}
  }
  \caption{A combined view of errors in RAG-Vanilla (failure rate \% of total queries).}
  \label{fig:combined_failures}
\end{figure}


\paragraph{Error analysis} Although the retrieval-based approach demonstrates significant improvements in accuracy, a closer examination of incorrect cases reveals important insights into its limitations. We analyze the error rates, distinguishing between errors occurring during the embedding retrieval stage and those during the language model’s response generation. The errors are categorized into two types:
\begin{itemize}
\item {Retrieval error}: The correct sentence is not included among the top-$k$ retrieved segments, indicating that the embedding model failed to identify the relevant content.
\item {LLM error}: The language model fails to extract or correctly reason about the target information, even when the relevant sentence is retrieved.
\end{itemize}

Error rates for each category are calculated as the percentage of total queries that result in that specific type of failure.

Figure~\ref{fig:failure_rates} shows the distribution of {retrieval errors} (red) versus {LLM errors} (blue) for GPT-4o-mini and Llama 3.2 90b using RAG-Vanilla. Although retrieval errors remain relatively low (4.6\% for both models), the LLM error rates are quite substantial for GPT-4o-mini (63.4\%) and Llama 3.2 90b (12.7\%), indicating that while RAG-Vanilla effectively narrows the context, the language model itself continues to struggle with multi-target reasoning. We also evaluated how increasing $k$ (the number of retrieved sentences) influences retrieval errors, as illustrated in Figure~\ref{fig:sentence_cap_analysis}. Although a larger $k$ reduces the likelihood of missing relevant information, it introduces additional distractors that can complicate the model’s reasoning process.

Based on these results, we identify several key factors contributing to LLM failures:
\begin{itemize}
    \item {Increased distractors}: Multiple similar sentences can confuse the model’s reasoning process.
    \item {Inconsistent answer prioritization}: Determining which label is “largest” or “most relevant” can be nontrivial when multiple valid options exist.
    \item {Ambiguity in sentence ranking}: Even with successful retrieval, the model may incorrectly prioritize semantically similar sentences when generating its final response.
\end{itemize}

These findings underscore the need for more advanced methods—such as neurosymbolic approaches—to improve multi-target reasoning beyond the gains offered by simple context narrowing.

\subsection{NeuroSymbolic Reasoning (NSAR)}
Although RAG narrows down the context and alleviates many challenges inherent to long input passages, it alone does not guarantee robust multi-target reasoning. To address this shortcoming, we enhance our baseline RAG system with NSAR component. 

As baselines, besides \textit{RAG-Vanilla}, we evaluate three prompting-based methods—\textit{Chain-of-Thought} (CoT), \textit{ReAct}, and \textit{Self-Reflection}. We also experiment with a hybrid approach (\textit{NSAR+3}) that combines NSAR with all three prompting strategies (Chain-of-Thought, ReAct, Self-Reflection, NSAR, NSAR+3). 

\begin{figure}[!t]
\centering
\includegraphics[width=0.95\linewidth]{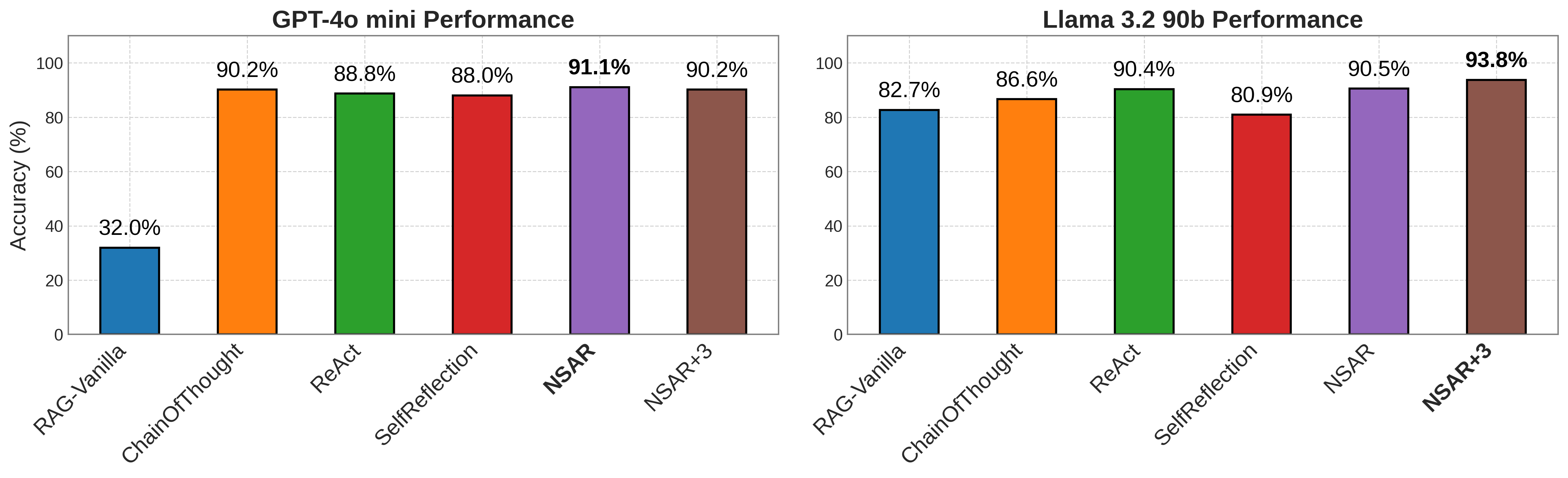}
    \caption{Overall accuracy of GPT-4o-mini (left) and Llama 3.2 90b (right) under different reasoning strategies ({RAG-Vanilla}, CoT, ReAct, Self-Reflection, {NSAR}, and a combined approach which combines NSAR with other reasoning methods (\textit{NSAR+3})).} \label{fig:neurosymbolic_comparison} \vspace{-0.7cm}
\end{figure}

As shown in Figure~\ref{fig:neurosymbolic_comparison}, the \textit{RAG-Vanilla} baseline lags behind in multi-target reasoning, confirming that narrowing the context alone does not suffice to fuse and compare multiple pieces of information. In contrast, our proposed approach {NSAR}  substantially improves accuracy by leveraging explicit symbolic extraction and Python-based reasoning. Moreover, combining NSAR with CoT, ReAct, and Self-Reflection (\textit{NSAR+3}) yields the highest accuracy overall. Notably, for GPT-4o-mini, \textit{NSAR} achieves 91.1\%, followed closely by \textit{NSAR+3} and \textit{CoT}, both at 90.2\%, whereas for Llama 3.2, \textit{NSAR+3} attains the highest performance at 93.8\%. These findings suggest that integrating explicit symbolic reasoning can fill critical gaps in retrieval-augmented generation, particularly for complex tasks that demand robust compositional inference.

\begin{figure}[!t]
  \centering
  \subfigure[GPT-4o mini.]{
    \includegraphics[width=0.45\textwidth]{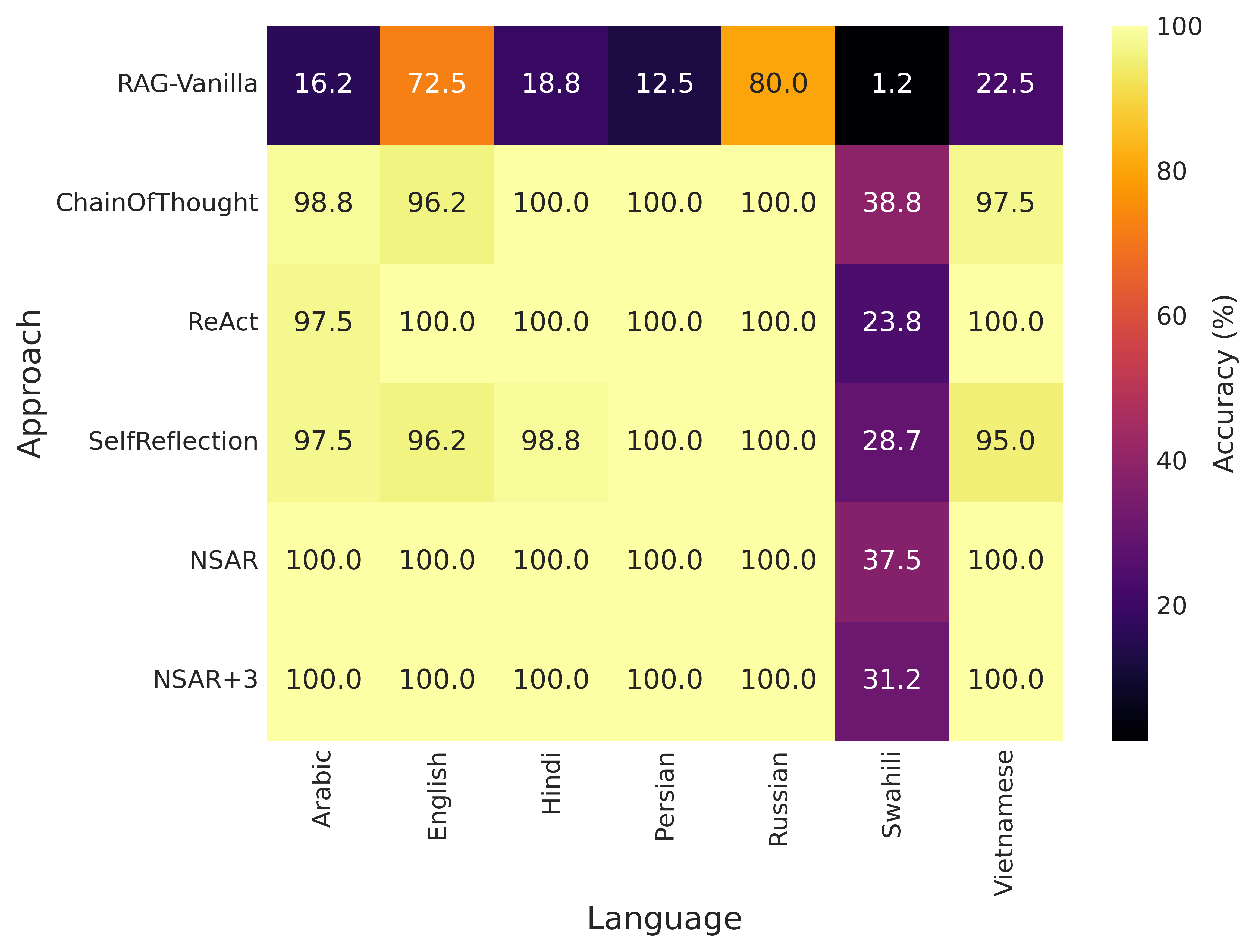}
    \label{fig:gpt4o_heatmap}
  }
  \hfill
  \subfigure[Llama 3.2 90b.]{
    \includegraphics[width=0.45\textwidth]{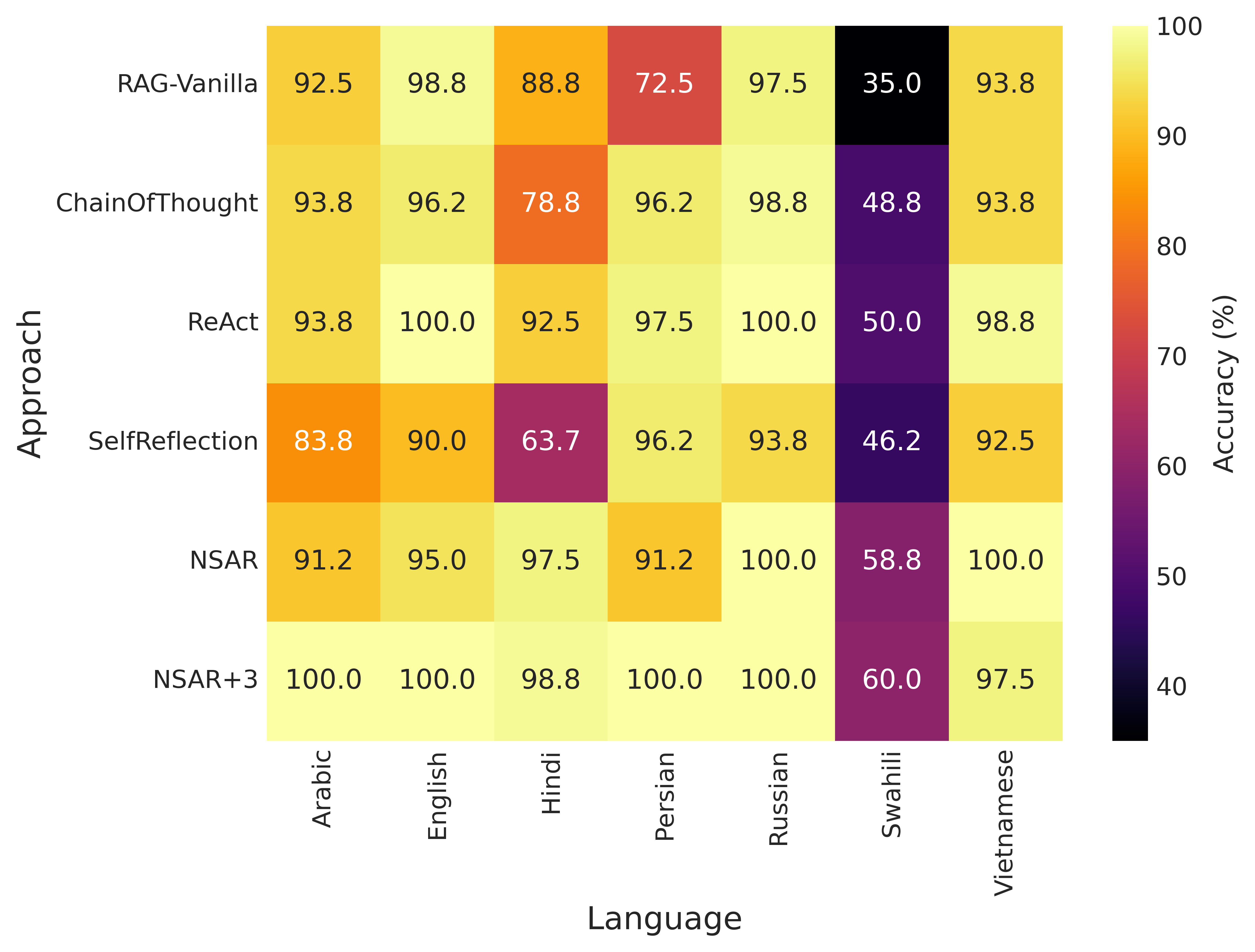}
    \label{fig:llama_heatmap}
  }
  \caption{
    Heatmaps illustrating the accuracy (\%) of different approaches (rows) across seven context languages (columns). 
    Darker cells indicate \textbf{lower} accuracy, while lighter cells indicate \textbf{higher} accuracy. 
  }
  \label{fig:heatmaps_comparison} \vspace{-0.7cm}
\end{figure}

\paragraph{Accuracy by context language}
Figures~\ref{fig:gpt4o_heatmap} and \ref{fig:llama_heatmap} provide a more granular perspective on how each approach performs across seven context languages. Each cell represents the accuracy (\%) of a particular approach–language pair, revealing where certain strategies excel or fall short.

\begin{figure}[!t]
    \centering
\includegraphics[width=0.95\linewidth]{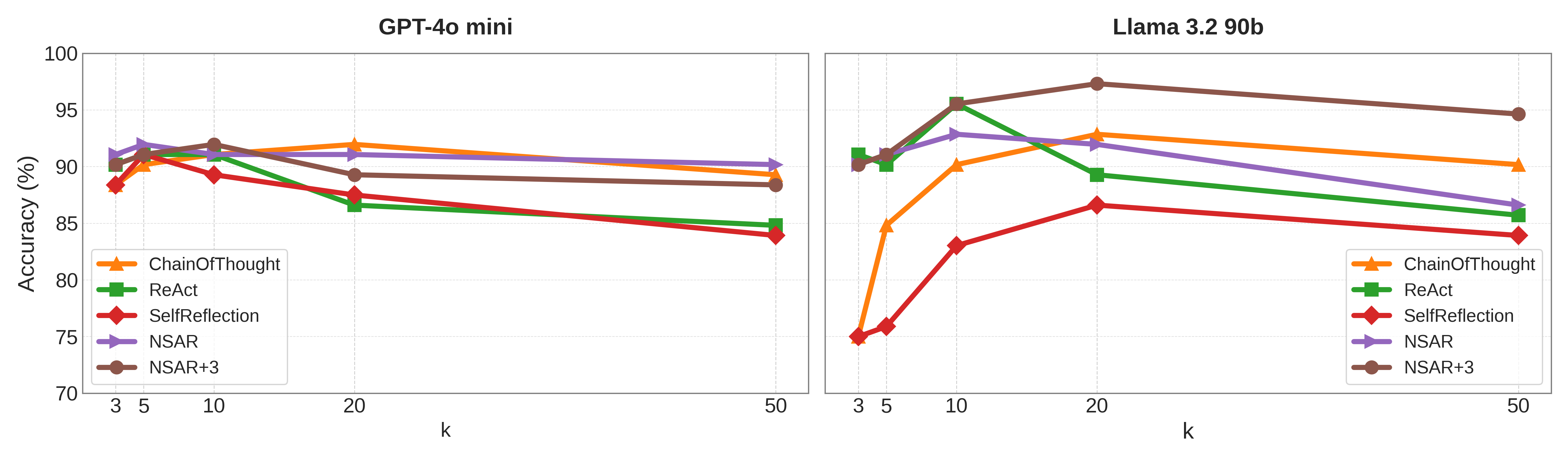}
    \caption{Accuracy versus $k$ (the number of retrieved sentences) for GPT-4o-mini (left) and Llama 3.2 (right). }
    \label{fig:accuracy_vs_k} \vspace{-0.7cm}
\end{figure}

Across both heatmaps, \textit{NSAR} and \textit{NSAR+3} maintain high accuracy in most languages, confirming the benefits of explicit symbolic reasoning for multi-target retrieval tasks. In contrast, baseline methods (\textit{RAG-Vanilla}) and single-step prompting strategies (Chain-of-Thought, ReAct, Self-Reflection) exhibit more variability and struggle in specific languages. Notably, languages such as Swahili and Arabic appear more challenging, yet neurosymbolic approaches still achieve competitive performance. These language-specific patterns underscore the importance of robust, compositional reasoning—particularly in cross-lingual or lower-resource settings.

\paragraph{Effect of $k$ on performance}
Figure~\ref{fig:accuracy_vs_k} shows how different reasoning strategies perform as we vary the number of retrieved sentences $k$ (3, 5, 10, 20, and 50) for GPT-4o-mini (left) and Llama 3.2 90b(right). Several trends are evident. First, at low values of $k$, all methods tend to have lower accuracy, likely due to the increased chance of missing key information during retrieval. As $k$ increases, accuracy generally improves up to a point. However, very large $k$ (e.g., 50) can introduce additional distractors, leading to a decline in performance. This aligns with our earlier observations that, while a broader retrieval scope reduces the risk of overlooking relevant facts, it can also complicate the model’s reasoning by introducing more non-essential content.

Comparing across models, Llama 3.2 shows more pronounced fluctuations as $k$ increases, suggesting it is more sensitive to context size and potential distractors. In contrast, GPT-4o-mini maintains relatively stable performance at intermediate $k$ values. Notably, NSAR+3 consistently outperforms purely neural prompting methods in Llama 3.2, whereas GPT-4o-mini exhibits closer competition among NSAR and Chain-of-Thought. Overall, these findings highlight the importance of carefully tuning $k$ to balance retrieval breadth and processing load, while also demonstrating that neurosymbolic reasoning can mitigate many of the challenges introduced by larger context windows.

\paragraph{Error analysis for NSAR and NSAR+3}

\begin{wrapfigure}{r}{0.45\textwidth}
    \centering
\includegraphics[width=\linewidth]{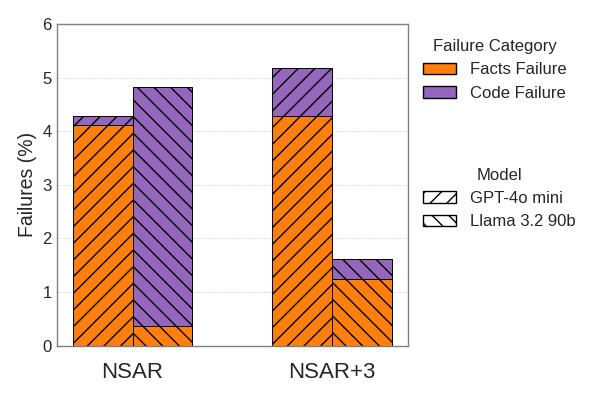}
    \caption{Failure rate (\% of total queries) for fact extraction vs. code generation errors under \emph{NSAR} and \emph{NSAR+3}.}
    \label{fig:nsar_failures} \vspace{-0.5cm}
\end{wrapfigure}

Figure~\ref{fig:nsar_failures} illustrates the distribution of error types for GPT-4o-mini and Llama~3.2 under the \emph{NSAR} and \emph{NSAR+3} methods, 
categorizing failures into two types:
\begin{itemize}
    \item {Facts}: The model retrieved the correct segments but failed to extract the target fact from the input.
    \item {Code}: Although the target fact was extracted correctly, the model produced incorrect or incomplete Python code, leading to an erroneous final answer.
\end{itemize}

The distribution of fact-extraction versus code-generation failures varies notably between the two models and across the two neurosymbolic methods. In \emph{NSAR} for GPT-4o-mini, most errors stem from fact extraction, whereas Llama~3.2 primarily struggles with code generation. This suggests that GPT-4o-mini’s neurosymbolic pipeline struggles in locating the correct sentences to extract facts, while Llama~3.2 successfully extracts facts but sometimes produces flawed Python code. Moving from \emph{NSAR} to \emph{NSAR+3} reverses this trend for Llama~3.2, significantly reducing code-generation failures but leading to more fact-extraction issues. Meanwhile, GPT-4o-mini exhibits a small increase in fact-extraction errors and a modest rise in code-generation errors when switching to \emph{NSAR+3}.

Overall, these results imply that while neurosymbolic reasoning substantially mitigates retrieval shortcomings, the balance between accurate fact extraction and correct code generation can shift depending on the underlying model and the specific prompting strategy.

\section{Conclusion}
We presented a neurosymbolic reasoning module (NSAR) to address the challenges of multi-target reasoning in long-context, cross-lingual settings. While RAG effectively narrows the context and reduces computational overhead, purely neural models often struggle to integrate and compare multiple pieces of information. NSAR overcomes these limitations by extracting symbolic facts and generating executable Python code, enabling transparent, verifiable reasoning. Our experiments across seven languages demonstrate that NSAR significantly outperforms both the vanilla RAG baseline and advanced prompting strategies (Chain-of-Thought, ReAct, and Self-Reflection). 
Future work will expand NSAR’s symbolic scope beyond numeric lookups—into graph queries, set operations, and constraint satisfaction—and evaluate on broader tasks to test generalizability of our reasoning pipeline.

\section*{Acknowledgments}
We thank the anonymous reviewers for their valuable feedback.

\bibliography{nesy2025-sample}

\begin{thebibliography}{25}
\providecommand{\natexlab}[1]{#1}
\providecommand{\url}[1]{\texttt{#1}}
\expandafter\ifx\csname urlstyle\endcsname\relax
  \providecommand{\doi}[1]{doi: #1}\else
  \providecommand{\doi}{doi: \begingroup \urlstyle{rm}\Url}\fi

\bibitem[Agrawal et~al.(2024)Agrawal, Dang, Nezhad, Pokharel, and Scheinberg]{agrawal2024evaluatingmultilinguallongcontextmodels}
Ameeta Agrawal, Andy Dang, Sina~Bagheri Nezhad, Rhitabrat Pokharel, and Russell Scheinberg.
\newblock Evaluating multilingual long-context models for retrieval and reasoning, 2024.
\newblock URL \url{https://arxiv.org/abs/2409.18006}.

\bibitem[Anthropic(2024)]{TheC3}
Anthropic.
\newblock The claude 3 model family: Opus, sonnet, haiku, 2024.
\newblock URL \url{https://www-cdn.anthropic.com/de8ba9b01c9ab7cbabf5c33b80b7bbc618857627/Model_Card_Claude_3.pdf}.

\bibitem[Chen et~al.(2024)Chen, Xiao, Zhang, Luo, Lian, and Liu]{chen-etal-2024-m3}
Jianlyu Chen, Shitao Xiao, Peitian Zhang, Kun Luo, Defu Lian, and Zheng Liu.
\newblock {M}3-embedding: Multi-linguality, multi-functionality, multi-granularity text embeddings through self-knowledge distillation.
\newblock In Lun-Wei Ku, Andre Martins, and Vivek Srikumar, editors, \emph{Findings of the Association for Computational Linguistics: ACL 2024}, pages 2318--2335, Bangkok, Thailand, August 2024. Association for Computational Linguistics.
\newblock \doi{10.18653/v1/2024.findings-acl.137}.
\newblock URL \url{https://aclanthology.org/2024.findings-acl.137}.

\bibitem[Fang et~al.(2024)Fang, Deng, Zhang, Shi, Chen, Pechenizkiy, and Wang]{fang2024large}
Meng Fang, Shilong Deng, Yudi Zhang, Zijing Shi, Ling Chen, Mykola Pechenizkiy, and Jun Wang.
\newblock Large language models are neurosymbolic reasoners.
\newblock In \emph{Proceedings of the AAAI conference on artificial intelligence}, volume~38, pages 17985--17993, 2024.

\bibitem[Gao et~al.(2023)Gao, Madaan, Zhou, Alon, Liu, Yang, Callan, and Neubig]{gao2023palprogramaidedlanguagemodels}
Luyu Gao, Aman Madaan, Shuyan Zhou, Uri Alon, Pengfei Liu, Yiming Yang, Jamie Callan, and Graham Neubig.
\newblock Pal: Program-aided language models, 2023.
\newblock URL \url{https://arxiv.org/abs/2211.10435}.

\bibitem[Hei et~al.(2024)Hei, Liu, Ou, Qiao, Jiao, Song, Tian, and Lin]{hei2024drragapplyingdynamicdocument}
Zijian Hei, Weiling Liu, Wenjie Ou, Juyi Qiao, Junming Jiao, Guowen Song, Ting Tian, and Yi~Lin.
\newblock Dr-rag: Applying dynamic document relevance to retrieval-augmented generation for question-answering, 2024.
\newblock URL \url{https://arxiv.org/abs/2406.07348}.

\bibitem[Hengle et~al.(2024)Hengle, Bajpai, Dan, and Chakraborty]{hengle2024multilingualneedlehaystackinvestigating}
Amey Hengle, Prasoon Bajpai, Soham Dan, and Tanmoy Chakraborty.
\newblock Multilingual needle in a haystack: Investigating long-context behavior of multilingual large language models, 2024.
\newblock URL \url{https://arxiv.org/abs/2408.10151}.

\bibitem[Huang et~al.(2023)Huang, Yu, and Allan]{Huang_2023}
Zhiqi Huang, Puxuan Yu, and James Allan.
\newblock Improving cross-lingual information retrieval on low-resource languages via optimal transport distillation.
\newblock In \emph{Proceedings of the Sixteenth ACM International Conference on Web Search and Data Mining}, WSDM ’23, page 1048–1056. ACM, February 2023.
\newblock \doi{10.1145/3539597.3570468}.
\newblock URL \url{http://dx.doi.org/10.1145/3539597.3570468}.

\bibitem[Jiang et~al.(2024)Jiang, Ma, and Chen]{jiang2024longragenhancingretrievalaugmentedgeneration}
Ziyan Jiang, Xueguang Ma, and Wenhu Chen.
\newblock Longrag: Enhancing retrieval-augmented generation with long-context llms, 2024.
\newblock URL \url{https://arxiv.org/abs/2406.15319}.

\bibitem[Kiss and Strunk(2006)]{10.1162/coli.2006.32.4.485}
Tibor Kiss and Jan Strunk.
\newblock Unsupervised multilingual sentence boundary detection.
\newblock \emph{Computational Linguistics}, 32\penalty0 (4):\penalty0 485--525, 12 2006.
\newblock ISSN 0891-2017.
\newblock \doi{10.1162/coli.2006.32.4.485}.
\newblock URL \url{https://doi.org/10.1162/coli.2006.32.4.485}.

\bibitem[Li et~al.(2024)Li, Haider, Luo, Agashe, and Callison-Burch]{li2024bordirlinesdatasetevaluatingcrosslingual}
Bryan Li, Samar Haider, Fiona Luo, Adwait Agashe, and Chris Callison-Burch.
\newblock Bordirlines: A dataset for evaluating cross-lingual retrieval-augmented generation, 2024.
\newblock URL \url{https://arxiv.org/abs/2410.01171}.

\bibitem[Limkonchotiwat et~al.(2024)Limkonchotiwat, Ponwitayarat, Lowphansirikul, Manakul, Udomcharoenchaikit, Chuangsuwanich, and Nutanong]{limkonchotiwat-etal-2024-mccrolin}
Peerat Limkonchotiwat, Wuttikorn Ponwitayarat, Lalita Lowphansirikul, Potsawee Manakul, Can Udomcharoenchaikit, Ekapol Chuangsuwanich, and Sarana Nutanong.
\newblock {M}c{C}rolin: Multi-consistency cross-lingual training for retrieval question answering.
\newblock In Yaser Al-Onaizan, Mohit Bansal, and Yun-Nung Chen, editors, \emph{Findings of the Association for Computational Linguistics: EMNLP 2024}, pages 2780--2793, Miami, Florida, USA, November 2024. Association for Computational Linguistics.
\newblock URL \url{https://aclanthology.org/2024.findings-emnlp.157}.

\bibitem[Lin et~al.(2024)Lin, Martins, and Schütze]{lin2024xamplerlearningretrievecrosslingual}
Peiqin Lin, André F.~T. Martins, and Hinrich Schütze.
\newblock Xampler: Learning to retrieve cross-lingual in-context examples, 2024.
\newblock URL \url{https://arxiv.org/abs/2405.05116}.

\bibitem[Litschko et~al.(2022)Litschko, Vuli{\'c}, and Glava{\v{s}}]{litschko-etal-2022-parameter}
Robert Litschko, Ivan Vuli{\'c}, and Goran Glava{\v{s}}.
\newblock Parameter-efficient neural reranking for cross-lingual and multilingual retrieval.
\newblock In Nicoletta Calzolari, Chu-Ren Huang, Hansaem Kim, James Pustejovsky, Leo Wanner, Key-Sun Choi, Pum-Mo Ryu, Hsin-Hsi Chen, Lucia Donatelli, Heng Ji, Sadao Kurohashi, Patrizia Paggio, Nianwen Xue, Seokhwan Kim, Younggyun Hahm, Zhong He, Tony~Kyungil Lee, Enrico Santus, Francis Bond, and Seung-Hoon Na, editors, \emph{Proceedings of the 29th International Conference on Computational Linguistics}, pages 1071--1082, Gyeongju, Republic of Korea, October 2022. International Committee on Computational Linguistics.
\newblock URL \url{https://aclanthology.org/2022.coling-1.90}.

\bibitem[Liu et~al.(2023)Liu, Lin, Hewitt, Paranjape, Bevilacqua, Petroni, and Liang]{liu2023lostmiddlelanguagemodels}
Nelson~F. Liu, Kevin Lin, John Hewitt, Ashwin Paranjape, Michele Bevilacqua, Fabio Petroni, and Percy Liang.
\newblock Lost in the middle: How language models use long contexts, 2023.
\newblock URL \url{https://arxiv.org/abs/2307.03172}.

\bibitem[Olausson et~al.(2023)Olausson, Gu, Lipkin, Zhang, Solar-Lezama, Tenenbaum, and Levy]{olausson-etal-2023-linc}
Theo Olausson, Alex Gu, Ben Lipkin, Cedegao Zhang, Armando Solar-Lezama, Joshua Tenenbaum, and Roger Levy.
\newblock {LINC}: A neurosymbolic approach for logical reasoning by combining language models with first-order logic provers.
\newblock In Houda Bouamor, Juan Pino, and Kalika Bali, editors, \emph{Proceedings of the 2023 Conference on Empirical Methods in Natural Language Processing}, pages 5153--5176, Singapore, December 2023. Association for Computational Linguistics.
\newblock \doi{10.18653/v1/2023.emnlp-main.313}.
\newblock URL \url{https://aclanthology.org/2023.emnlp-main.313/}.

\bibitem[OpenAI et~al.(2024)OpenAI, :, Hurst, Lerer, Goucher, Perelman, Ramesh, Clark, Ostrow, Welihinda, Hayes, Radford, Mądry, Baker-Whitcomb, Beutel, Borzunov, Carney, Chow, Kirillov, Nichol, Paino, Renzin, Passos, Kirillov, Christakis, Conneau, Kamali, Jabri, Moyer, Tam, Crookes, Tootoochian, Tootoonchian, Kumar, Vallone, Karpathy, Braunstein, Cann, Codispoti, Galu, Kondrich, Tulloch, Mishchenko, Baek, Jiang, Pelisse, Woodford, Gosalia, Dhar, Pantuliano, Nayak, Oliver, Zoph, Ghorbani, Leimberger, Rossen, Sokolowsky, Wang, Zweig, Hoover, Samic, McGrew, Spero, Giertler, Cheng, Lightcap, Walkin, Quinn, Guarraci, Hsu, Kellogg, Eastman, Lugaresi, Wainwright, Bassin, Hudson, Chu, Nelson, Li, Shern, Conger, Barette, Voss, Ding, Lu, Zhang, Beaumont, Hallacy, Koch, Gibson, Kim, Choi, McLeavey, Hesse, Fischer, Winter, Czarnecki, Jarvis, Wei, Koumouzelis, Sherburn, Kappler, Levin, Levy, Carr, Farhi, Mely, Robinson, Sasaki, Jin, Valladares, Tsipras, Li, Nguyen, Findlay, Oiwoh, Wong, Asdar, Proehl, Yang, Antonow,
  Kramer, Peterson, Sigler, Wallace, Brevdo, Mays, Khorasani, Such, Raso, Zhang, von Lohmann, Sulit, Goh, Oden, Salmon, Starace, Brockman, Salman, Bao, Hu, Wong, Wang, Schmidt, Whitney, Jun, Kirchner, de~Oliveira~Pinto, Ren, Chang, Chung, Kivlichan, O'Connell, O'Connell, Osband, Silber, Sohl, Okuyucu, Lan, Kostrikov, Sutskever, Kanitscheider, Gulrajani, Coxon, Menick, Pachocki, Aung, Betker, Crooks, Lennon, Kiros, Leike, Park, Kwon, Phang, Teplitz, Wei, Wolfe, Chen, Harris, Varavva, Lee, Shieh, Lin, Yu, Weng, Tang, Yu, Jang, Candela, Beutler, Landers, Parish, Heidecke, Schulman, Lachman, McKay, Uesato, Ward, Kim, Huizinga, Sitkin, Kraaijeveld, Gross, Kaplan, Snyder, Achiam, Jiao, Lee, Zhuang, Harriman, Fricke, Hayashi, Singhal, Shi, Karthik, Wood, Rimbach, Hsu, Nguyen, Gu-Lemberg, Button, Liu, Howe, Muthukumar, Luther, Ahmad, Kai, Itow, Workman, Pathak, Chen, Jing, Guy, Fedus, Zhou, Mamitsuka, Weng, McCallum, Held, Ouyang, Feuvrier, Zhang, Kondraciuk, Kaiser, Hewitt, Metz, Doshi, Aflak, Simens, Boyd,
  Thompson, Dukhan, Chen, Gray, Hudnall, Zhang, Aljubeh, Litwin, Zeng, Johnson, Shetty, Gupta, Shah, Yatbaz, Yang, Zhong, Glaese, Chen, Janner, Lampe, Petrov, Wu, Wang, Fradin, Pokrass, Castro, de~Castro, Pavlov, Brundage, Wang, Khan, Murati, Bavarian, Lin, Yesildal, Soto, Gimelshein, Cone, Staudacher, Summers, LaFontaine, Chowdhury, Ryder, Stathas, Turley, Tezak, Felix, Kudige, Keskar, Deutsch, Bundick, Puckett, Nachum, Okelola, Boiko, Murk, Jaffe, Watkins, Godement, Campbell-Moore, Chao, McMillan, Belov, Su, Bak, Bakkum, Deng, Dolan, Hoeschele, Welinder, Tillet, Pronin, Tillet, Dhariwal, Yuan, Dias, Lim, Arora, Troll, Lin, Lopes, Puri, Miyara, Leike, Gaubert, Zamani, Wang, Donnelly, Honsby, Smith, Sahai, Ramchandani, Huet, Carmichael, Zellers, Chen, Chen, Nigmatullin, Cheu, Jain, Altman, Schoenholz, Toizer, Miserendino, Agarwal, Culver, Ethersmith, Gray, Grove, Metzger, Hermani, Jain, Zhao, Wu, Jomoto, Wu, Shuaiqi, Xia, Phene, Papay, Narayanan, Coffey, Lee, Hall, Balaji, Broda, Stramer, Xu, Gogineni,
  Christianson, Sanders, Patwardhan, Cunninghman, Degry, Dimson, Raoux, Shadwell, Zheng, Underwood, Markov, Sherbakov, Rubin, Stasi, Kaftan, Heywood, Peterson, Walters, Eloundou, Qi, Moeller, Monaco, Kuo, Fomenko, Chang, Zheng, Zhou, Manassra, Sheu, Zaremba, Patil, Qian, Kim, Cheng, Zhang, He, Zhang, Jin, Dai, and Malkov]{openai2024gpt4ocard}
OpenAI, :, Aaron Hurst, Adam Lerer, Adam~P. Goucher, Adam Perelman, Aditya Ramesh, Aidan Clark, AJ~Ostrow, Akila Welihinda, Alan Hayes, Alec Radford, Aleksander Mądry, Alex Baker-Whitcomb, Alex Beutel, Alex Borzunov, Alex Carney, Alex Chow, Alex Kirillov, Alex Nichol, Alex Paino, Alex Renzin, Alex~Tachard Passos, Alexander Kirillov, Alexi Christakis, Alexis Conneau, Ali Kamali, Allan Jabri, Allison Moyer, Allison Tam, Amadou Crookes, Amin Tootoochian, Amin Tootoonchian, Ananya Kumar, Andrea Vallone, Andrej Karpathy, Andrew Braunstein, Andrew Cann, Andrew Codispoti, Andrew Galu, Andrew Kondrich, Andrew Tulloch, Andrey Mishchenko, Angela Baek, Angela Jiang, Antoine Pelisse, Antonia Woodford, Anuj Gosalia, Arka Dhar, Ashley Pantuliano, Avi Nayak, Avital Oliver, Barret Zoph, Behrooz Ghorbani, Ben Leimberger, Ben Rossen, Ben Sokolowsky, Ben Wang, Benjamin Zweig, Beth Hoover, Blake Samic, Bob McGrew, Bobby Spero, Bogo Giertler, Bowen Cheng, Brad Lightcap, Brandon Walkin, Brendan Quinn, Brian Guarraci, Brian Hsu,
  Bright Kellogg, Brydon Eastman, Camillo Lugaresi, Carroll Wainwright, Cary Bassin, Cary Hudson, Casey Chu, Chad Nelson, Chak Li, Chan~Jun Shern, Channing Conger, Charlotte Barette, Chelsea Voss, Chen Ding, Cheng Lu, Chong Zhang, Chris Beaumont, Chris Hallacy, Chris Koch, Christian Gibson, Christina Kim, Christine Choi, Christine McLeavey, Christopher Hesse, Claudia Fischer, Clemens Winter, Coley Czarnecki, Colin Jarvis, Colin Wei, Constantin Koumouzelis, Dane Sherburn, Daniel Kappler, Daniel Levin, Daniel Levy, David Carr, David Farhi, David Mely, David Robinson, David Sasaki, Denny Jin, Dev Valladares, Dimitris Tsipras, Doug Li, Duc~Phong Nguyen, Duncan Findlay, Edede Oiwoh, Edmund Wong, Ehsan Asdar, Elizabeth Proehl, Elizabeth Yang, Eric Antonow, Eric Kramer, Eric Peterson, Eric Sigler, Eric Wallace, Eugene Brevdo, Evan Mays, Farzad Khorasani, Felipe~Petroski Such, Filippo Raso, Francis Zhang, Fred von Lohmann, Freddie Sulit, Gabriel Goh, Gene Oden, Geoff Salmon, Giulio Starace, Greg Brockman, Hadi
  Salman, Haiming Bao, Haitang Hu, Hannah Wong, Haoyu Wang, Heather Schmidt, Heather Whitney, Heewoo Jun, Hendrik Kirchner, Henrique~Ponde de~Oliveira~Pinto, Hongyu Ren, Huiwen Chang, Hyung~Won Chung, Ian Kivlichan, Ian O'Connell, Ian O'Connell, Ian Osband, Ian Silber, Ian Sohl, Ibrahim Okuyucu, Ikai Lan, Ilya Kostrikov, Ilya Sutskever, Ingmar Kanitscheider, Ishaan Gulrajani, Jacob Coxon, Jacob Menick, Jakub Pachocki, James Aung, James Betker, James Crooks, James Lennon, Jamie Kiros, Jan Leike, Jane Park, Jason Kwon, Jason Phang, Jason Teplitz, Jason Wei, Jason Wolfe, Jay Chen, Jeff Harris, Jenia Varavva, Jessica~Gan Lee, Jessica Shieh, Ji~Lin, Jiahui Yu, Jiayi Weng, Jie Tang, Jieqi Yu, Joanne Jang, Joaquin~Quinonero Candela, Joe Beutler, Joe Landers, Joel Parish, Johannes Heidecke, John Schulman, Jonathan Lachman, Jonathan McKay, Jonathan Uesato, Jonathan Ward, Jong~Wook Kim, Joost Huizinga, Jordan Sitkin, Jos Kraaijeveld, Josh Gross, Josh Kaplan, Josh Snyder, Joshua Achiam, Joy Jiao, Joyce Lee, Juntang
  Zhuang, Justyn Harriman, Kai Fricke, Kai Hayashi, Karan Singhal, Katy Shi, Kavin Karthik, Kayla Wood, Kendra Rimbach, Kenny Hsu, Kenny Nguyen, Keren Gu-Lemberg, Kevin Button, Kevin Liu, Kiel Howe, Krithika Muthukumar, Kyle Luther, Lama Ahmad, Larry Kai, Lauren Itow, Lauren Workman, Leher Pathak, Leo Chen, Li~Jing, Lia Guy, Liam Fedus, Liang Zhou, Lien Mamitsuka, Lilian Weng, Lindsay McCallum, Lindsey Held, Long Ouyang, Louis Feuvrier, Lu~Zhang, Lukas Kondraciuk, Lukasz Kaiser, Luke Hewitt, Luke Metz, Lyric Doshi, Mada Aflak, Maddie Simens, Madelaine Boyd, Madeleine Thompson, Marat Dukhan, Mark Chen, Mark Gray, Mark Hudnall, Marvin Zhang, Marwan Aljubeh, Mateusz Litwin, Matthew Zeng, Max Johnson, Maya Shetty, Mayank Gupta, Meghan Shah, Mehmet Yatbaz, Meng~Jia Yang, Mengchao Zhong, Mia Glaese, Mianna Chen, Michael Janner, Michael Lampe, Michael Petrov, Michael Wu, Michele Wang, Michelle Fradin, Michelle Pokrass, Miguel Castro, Miguel Oom~Temudo de~Castro, Mikhail Pavlov, Miles Brundage, Miles Wang, Minal
  Khan, Mira Murati, Mo~Bavarian, Molly Lin, Murat Yesildal, Nacho Soto, Natalia Gimelshein, Natalie Cone, Natalie Staudacher, Natalie Summers, Natan LaFontaine, Neil Chowdhury, Nick Ryder, Nick Stathas, Nick Turley, Nik Tezak, Niko Felix, Nithanth Kudige, Nitish Keskar, Noah Deutsch, Noel Bundick, Nora Puckett, Ofir Nachum, Ola Okelola, Oleg Boiko, Oleg Murk, Oliver Jaffe, Olivia Watkins, Olivier Godement, Owen Campbell-Moore, Patrick Chao, Paul McMillan, Pavel Belov, Peng Su, Peter Bak, Peter Bakkum, Peter Deng, Peter Dolan, Peter Hoeschele, Peter Welinder, Phil Tillet, Philip Pronin, Philippe Tillet, Prafulla Dhariwal, Qiming Yuan, Rachel Dias, Rachel Lim, Rahul Arora, Rajan Troll, Randall Lin, Rapha~Gontijo Lopes, Raul Puri, Reah Miyara, Reimar Leike, Renaud Gaubert, Reza Zamani, Ricky Wang, Rob Donnelly, Rob Honsby, Rocky Smith, Rohan Sahai, Rohit Ramchandani, Romain Huet, Rory Carmichael, Rowan Zellers, Roy Chen, Ruby Chen, Ruslan Nigmatullin, Ryan Cheu, Saachi Jain, Sam Altman, Sam Schoenholz, Sam
  Toizer, Samuel Miserendino, Sandhini Agarwal, Sara Culver, Scott Ethersmith, Scott Gray, Sean Grove, Sean Metzger, Shamez Hermani, Shantanu Jain, Shengjia Zhao, Sherwin Wu, Shino Jomoto, Shirong Wu, Shuaiqi, Xia, Sonia Phene, Spencer Papay, Srinivas Narayanan, Steve Coffey, Steve Lee, Stewart Hall, Suchir Balaji, Tal Broda, Tal Stramer, Tao Xu, Tarun Gogineni, Taya Christianson, Ted Sanders, Tejal Patwardhan, Thomas Cunninghman, Thomas Degry, Thomas Dimson, Thomas Raoux, Thomas Shadwell, Tianhao Zheng, Todd Underwood, Todor Markov, Toki Sherbakov, Tom Rubin, Tom Stasi, Tomer Kaftan, Tristan Heywood, Troy Peterson, Tyce Walters, Tyna Eloundou, Valerie Qi, Veit Moeller, Vinnie Monaco, Vishal Kuo, Vlad Fomenko, Wayne Chang, Weiyi Zheng, Wenda Zhou, Wesam Manassra, Will Sheu, Wojciech Zaremba, Yash Patil, Yilei Qian, Yongjik Kim, Youlong Cheng, Yu~Zhang, Yuchen He, Yuchen Zhang, Yujia Jin, Yunxing Dai, and Yury Malkov.
\newblock Gpt-4o system card, 2024.
\newblock URL \url{https://arxiv.org/abs/2410.21276}.

\bibitem[Renze and Guven(2024)]{renze2024selfreflectionllmagentseffects}
Matthew Renze and Erhan Guven.
\newblock Self-reflection in llm agents: Effects on problem-solving performance, 2024.
\newblock URL \url{https://arxiv.org/abs/2405.06682}.

\bibitem[Team(2024)]{geminiteam2024gemini15unlockingmultimodal}
Gemini Team.
\newblock Gemini 1.5: Unlocking multimodal understanding across millions of tokens of context, 2024.
\newblock URL \url{https://arxiv.org/abs/2403.05530}.

\bibitem[Touvron et~al.(2023)Touvron, Martin, Stone, Albert, Almahairi, Babaei, Bashlykov, Batra, Bhargava, Bhosale, Bikel, Blecher, Ferrer, Chen, Cucurull, Esiobu, Fernandes, Fu, Fu, Fuller, Gao, Goswami, Goyal, Hartshorn, Hosseini, Hou, Inan, Kardas, Kerkez, Khabsa, Kloumann, Korenev, Koura, Lachaux, Lavril, Lee, Liskovich, Lu, Mao, Martinet, Mihaylov, Mishra, Molybog, Nie, Poulton, Reizenstein, Rungta, Saladi, Schelten, Silva, Smith, Subramanian, Tan, Tang, Taylor, Williams, Kuan, Xu, Yan, Zarov, Zhang, Fan, Kambadur, Narang, Rodriguez, Stojnic, Edunov, and Scialom]{touvron2023llama2openfoundation}
Hugo Touvron, Louis Martin, Kevin Stone, Peter Albert, Amjad Almahairi, Yasmine Babaei, Nikolay Bashlykov, Soumya Batra, Prajjwal Bhargava, Shruti Bhosale, Dan Bikel, Lukas Blecher, Cristian~Canton Ferrer, Moya Chen, Guillem Cucurull, David Esiobu, Jude Fernandes, Jeremy Fu, Wenyin Fu, Brian Fuller, Cynthia Gao, Vedanuj Goswami, Naman Goyal, Anthony Hartshorn, Saghar Hosseini, Rui Hou, Hakan Inan, Marcin Kardas, Viktor Kerkez, Madian Khabsa, Isabel Kloumann, Artem Korenev, Punit~Singh Koura, Marie-Anne Lachaux, Thibaut Lavril, Jenya Lee, Diana Liskovich, Yinghai Lu, Yuning Mao, Xavier Martinet, Todor Mihaylov, Pushkar Mishra, Igor Molybog, Yixin Nie, Andrew Poulton, Jeremy Reizenstein, Rashi Rungta, Kalyan Saladi, Alan Schelten, Ruan Silva, Eric~Michael Smith, Ranjan Subramanian, Xiaoqing~Ellen Tan, Binh Tang, Ross Taylor, Adina Williams, Jian~Xiang Kuan, Puxin Xu, Zheng Yan, Iliyan Zarov, Yuchen Zhang, Angela Fan, Melanie Kambadur, Sharan Narang, Aurelien Rodriguez, Robert Stojnic, Sergey Edunov, and Thomas
  Scialom.
\newblock Llama 2: Open foundation and fine-tuned chat models, 2023.
\newblock URL \url{https://arxiv.org/abs/2307.09288}.

\bibitem[Wei et~al.(2022)Wei, Wang, Schuurmans, Bosma, Xia, Chi, Le, Zhou, et~al.]{wei2022chain}
Jason Wei, Xuezhi Wang, Dale Schuurmans, Maarten Bosma, Fei Xia, Ed~Chi, Quoc~V Le, Denny Zhou, et~al.
\newblock Chain-of-thought prompting elicits reasoning in large language models.
\newblock \emph{Advances in neural information processing systems}, 35:\penalty0 24824--24837, 2022.

\bibitem[Xu et~al.(2024)Xu, Ping, Wu, McAfee, Zhu, Liu, Subramanian, Bakhturina, Shoeybi, and Catanzaro]{xu2024retrieval}
Peng Xu, Wei Ping, Xianchao Wu, Lawrence McAfee, Chen Zhu, Zihan Liu, Sandeep Subramanian, Evelina Bakhturina, Mohammad Shoeybi, and Bryan Catanzaro.
\newblock Retrieval meets long context large language models.
\newblock In \emph{The Twelfth International Conference on Learning Representations}, 2024.
\newblock URL \url{https://openreview.net/forum?id=xw5nxFWMlo}.

\bibitem[Yao et~al.(2023{\natexlab{a}})Yao, Yu, Zhao, Shafran, Griffiths, Cao, and Narasimhan]{yao2023tree}
Shunyu Yao, Dian Yu, Jeffrey Zhao, Izhak Shafran, Tom Griffiths, Yuan Cao, and Karthik Narasimhan.
\newblock Tree of thoughts: Deliberate problem solving with large language models.
\newblock \emph{Advances in neural information processing systems}, 36:\penalty0 11809--11822, 2023{\natexlab{a}}.

\bibitem[Yao et~al.(2023{\natexlab{b}})Yao, Zhao, Yu, Du, Shafran, Narasimhan, and Cao]{yao2023react}
Shunyu Yao, Jeffrey Zhao, Dian Yu, Nan Du, Izhak Shafran, Karthik~R Narasimhan, and Yuan Cao.
\newblock React: Synergizing reasoning and acting in language models.
\newblock In \emph{The Eleventh International Conference on Learning Representations}, 2023{\natexlab{b}}.
\newblock URL \url{https://openreview.net/forum?id=WE_vluYUL-X}.

\bibitem[Zhu et~al.(2024)Zhu, Yuan, Wang, Liu, Liu, Deng, Chen, Liu, Dou, and Wen]{zhu2024largelanguagemodelsinformation}
Yutao Zhu, Huaying Yuan, Shuting Wang, Jiongnan Liu, Wenhan Liu, Chenlong Deng, Haonan Chen, Zheng Liu, Zhicheng Dou, and Ji-Rong Wen.
\newblock Large language models for information retrieval: A survey, 2024.
\newblock URL \url{https://arxiv.org/abs/2308.07107}.

\end{thebibliography}

\appendix
\section{Prompt and Needle Templates}

This appendix presents the templates and data used in our experiments. For the corresponding versions of Needles in other languages (Swahili, Vietnamese, Persian, Hindi, Arabic, and Russian), please refer to the anonymous repository \footnote{\url{https://anonymous.4open.science/r/prompts-template-D729}}.

\subsection{Needle Template}
\label{sec:needels}
\begin{verbatim}
"The special magic {city} number is {number}."
\end{verbatim}
\noindent Here, \texttt{number} represents a randomly generated 7-digit number, and \texttt{city} is selected at random from the list below (with the city names translated into the context language):

\begin{multicols}{3}
\begin{itemize}
    \item Chicago
    \item Yangon
    \item Antananarivo
    \item Colombo
    \item Almaty
    \item Sydney
    \item Mexico City
    \item Seattle
    \item Lagos
    \item Amsterdam
    \item Belgrade
    \item Cairo
    \item Baghdad
    \item Damascus
    \item Kigali
    \item Dakar
    \item Sofia
    \item Victoria
    \item Tashkent
    \item Mumbai
    \item Barcelona
    \item Amman
    \item Toronto
\end{itemize}
\end{multicols}

\subsection{Prompt Templates}
\label{sec:prompts}

This appendix presents the complete set of prompt templates used in our experiments, each designed to elicit concise and direct answers from a large language model (LLM). All prompts share the convention that contextual information is enclosed between \texttt{\#CONTEXT} and \texttt{\#ENDCONTEXT}, followed by a specific question. 

\begin{itemize}
    \item \textbf{Vanilla Prompts}: Adapted from \citet{agrawal2024evaluatingmultilinguallongcontextmodels}, provide a baseline retrieval-focused setup without any specialized reasoning strategy.
    \item \textbf{Chain-of-Thought (CoT)}: Encourages the LLM to articulate its reasoning steps in natural language before arriving at a final answer, aiming to improve transparency and correctness.
    \item \textbf{ReAct}: Combines “thinking” (Thought) and “acting” (Action) phases, prompting the LLM to explicitly delineate its internal reasoning and the subsequent steps taken to reach an answer.
    \item \textbf{Self-Reflection}: Prompts the model to not only provide a step-by-step solution but also critically review its reasoning for potential errors, thus refining the final answer.
    \item \textbf{NSAR}: Introduces a neurosymbolic prompt that extracts symbolic facts and generates executable Python code to produce a deterministic, verifiable answer.
    \item \textbf{NSAR+3}: A hybrid method that combines neurosymbolic reasoning with Chain-of-Thought, ReAct, and Self-Reflection, offering a comprehensive approach to both interpretability and robust logic.
\end{itemize}

Below, we detail each prompt template:

\begin{tcolorbox}[title=Vanilla Prompt Template, breakable, enhanced]
You are a helpful AI bot that answers questions for a user. Keep your response short and direct.

\#CONTEXT

\{text\}

\#ENDCONTEXT

\#QUESTION

What is the largest special magic number? 
Don't give information outside the document or repeat your findings.
If the information is not available in the context, respond UNANSWERABLE.
\end{tcolorbox}

\begin{tcolorbox}[title=Chain-of-Thought Prompt Template, breakable, enhanced]
You are a helpful assistant. Below is a context and a question. Please think through the problem step by step, detailing your reasoning process thoroughly before providing your final answer.

\#CONTEXT

\{text\}

\#ENDCONTEXT

\#QUESTION

What is the largest special magic number?
Please explain your reasoning process step by step before providing the final answer.
\end{tcolorbox}

\begin{tcolorbox}[title=ReAct Prompt Template, breakable, enhanced]
You are a helpful assistant. Below is a context and a question. For this question, please follow these steps:

1. Provide your thought process, prefixed with "Thought:".

2. Describe the action you would take, prefixed with "Action:".

3. Finally, state the final answer, prefixed with "Final Answer:".

\#CONTEXT

\{text\}

\#ENDCONTEXT

\#QUESTION

What is the largest special magic number?
\end{tcolorbox}

\begin{tcolorbox}[title=Self-Reflection Prompt Template, breakable, enhanced]
You are a helpful assistant. Below is a context and a question. For the given question, please:

1. Provide a detailed, step-by-step explanation of your reasoning.

2. Critically review your reasoning to ensure it is sound.

3. Finally, state your final answer.

\#CONTEXT

\{text\}

\#ENDCONTEXT

\#QUESTION

What is the largest special magic number?
\end{tcolorbox}

\begin{tcolorbox}[title=NSAR Prompt Template, breakable, enhanced]
You are a helpful assistant that employs a neurosymbolic method. Given the following context and question, please follow these steps:

1. Extract all relevant facts from the context and represent them as symbolic facts using the format \texttt{FACT(entity, attribute, value)}.

2. Generate executable Python code that uses the extracted symbolic facts to compute the final answer.

3. Finally, output only the final answer.

\#CONTEXT

\{text\}

\#ENDCONTEXT

\#QUESTION

What is the largest special magic number?
\end{tcolorbox}

\begin{tcolorbox}[title=NSAR+3 Prompt Template, breakable, enhanced]
You are a helpful assistant that employs a neurosymbolic method combining chain-of-thought, ReAct, and self-reflection. Given the following context and question, please follow these steps:

1. Extract all relevant facts from the context and represent them as symbolic facts using the format \texttt{FACT(entity, attribute, value)}.

2. Provide a detailed, step-by-step chain-of-thought explanation of your reasoning.

3. Describe the action you would take (e.g., generating and executing Python code) to compute the answer.

4. Generate executable Python code that uses the extracted symbolic facts to compute the final answer.

5. Reflect on your reasoning process to verify its soundness.

6. Finally, output only the final answer.

\#CONTEXT

\{text\}

\#ENDCONTEXT

\#QUESTION

What is the largest special magic number?
\end{tcolorbox}

Each of these templates is tailored to evaluate different aspects of the LLM’s reasoning and retrieval capabilities. The \textbf{Vanilla} prompts test basic and multi-target retrieval scenarios, while Chain-of-Thought, ReAct, and Self-Reflection aim to improve intermediate reasoning steps. Meanwhile, \textbf{NSAR} introduces an explicit neurosymbolic approach for verifiable multi-target logic, and \textbf{NSAR+3} extends this by integrating advanced prompting strategies for even more robust reasoning.

\end{document}